\documentclass[10pt,twocolumn,letterpaper]{article}

\usepackage[pagenumbers]{cvpr} 

\usepackage[dvipsnames]{xcolor}

\usepackage{algorithm, algorithmic, setspace}
\usepackage{booktabs,multirow,longtable,makecell,graphicx}
\usepackage{stfloats}

\definecolor{cvprblue}{rgb}{0.21,0.49,0.74}
\usepackage[pagebackref,breaklinks,colorlinks,citecolor=cvprblue]{hyperref}

\def\paperID{9787} 
\def\confName{CVPR}
\def\confYear{2024}

\title{UDiffText: A Unified Framework for High-quality Text Synthesis in Arbitrary Images via Character-aware Diffusion Models}

\author{Yiming Zhao,~~~Zhouhui Lian\\
Wangxuan Institute of Computer Technology\\
Peking University, Beijing, China\\
{\tt\small \{zhaoym, lianzhouhui\}@pku.edu.cn}
}

\begin{document}
\twocolumn[{%
\maketitle
\begin{figure}[H]
\hsize=\textwidth
\centering
\vspace{-0.8cm}
\includegraphics[width=\textwidth]{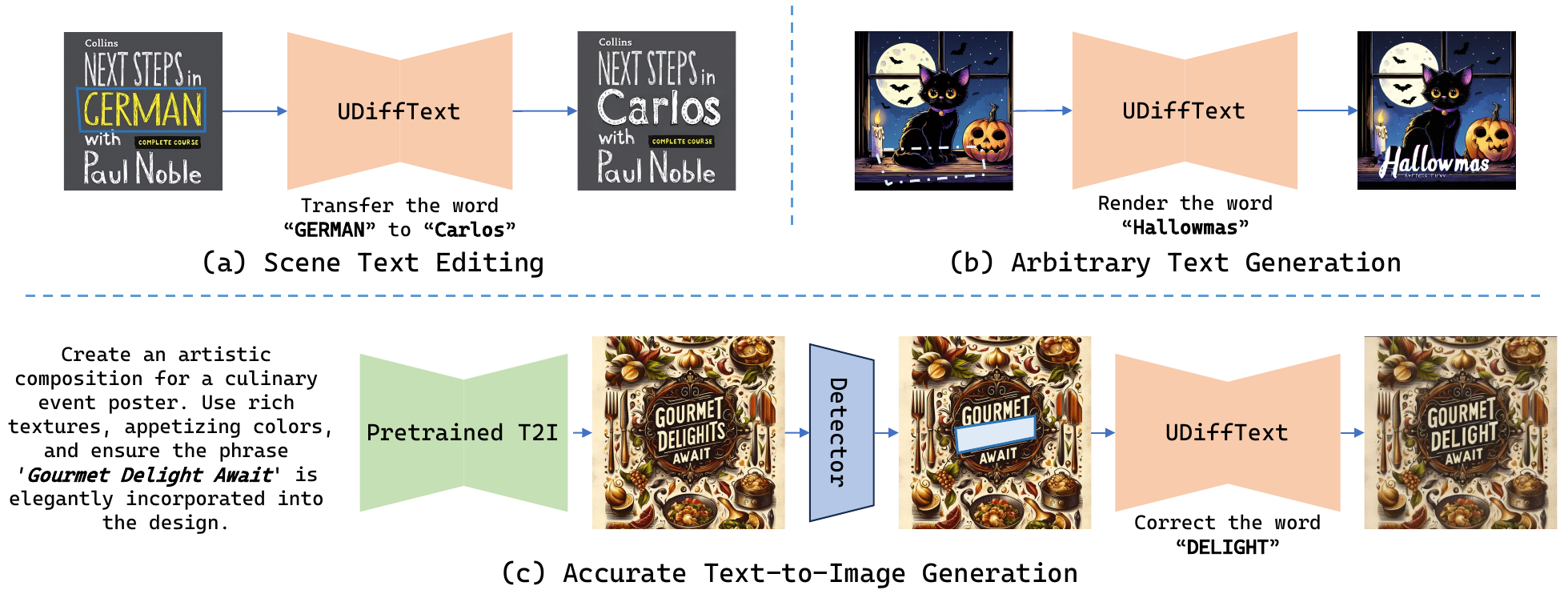}
\caption{The proposed UDiffText is capable of synthesizing accurate and harmonious text in either synthetic or real-word images, thus can be applied to tasks like scene text editing (a), arbitrary text generation (b) and accurate T2I generation (c).}
\label{fig:teaser}
\end{figure}
}] 
\begin{abstract}
    
Text-to-Image (T2I) generation methods based on diffusion model have garnered significant attention in the last few years. Although these image synthesis methods produce visually appealing results, they frequently exhibit spelling errors when rendering text within the generated images. Such errors manifest as missing, incorrect or extraneous characters, thereby severely constraining the performance of text image generation based on diffusion models. To address the aforementioned issue, this paper proposes a novel approach for text image generation, utilizing a pre-trained diffusion model (i.e., Stable Diffusion~\cite{rombach2022high}). Our approach involves the design and training of a light-weight character-level text encoder, which replaces the original CLIP encoder and provides more robust text embeddings as conditional guidance. Then, we  fine-tune the diffusion model using a large-scale dataset, incorporating local attention control under the supervision of character-level segmentation maps. Finally, by employing an inference stage refinement process, we achieve a notably high sequence accuracy when synthesizing text in arbitrarily given images. Both qualitative and quantitative results demonstrate the superiority of our method to the state of the art. Furthermore, we showcase several potential applications of the proposed UDiffText, including text-centric image synthesis, scene text editing, etc. Code and model will be available at \href{https://github.com/ZYM-PKU/UDiffText}{https://github.com/ZYM-PKU/UDiffText}.

\vspace{-0.4cm}
\end{abstract}
    
\section{Introduction}
\label{sec:intro}

Since the proposal of denoising diffusion probability model (DDPM)~\cite{ho2020denoising}, it has shown great potential in the field of image generation. In comparison with traditional generative adversarial networks (GANs)~\cite{goodfellow2020generative}, this kind of hidden-variable probabilistic graphical model has significant advantages, which are specifically reflected in its simple optimization objectives and clear iterative definition of the generation process. Especially, it does not suffer the problem of loss convergence difficulty when the model parameters are expanded. With the evolution of multimodal approaches, the integration of textual guidance into the diffusion model has given rise to large T2I generation models~\cite{ramesh2022hierarchical, saharia2022photorealistic, rombach2022high, Betker2023dalle3, podell2023sdxl}. The majority of these models have a substantial number of parameters and are trained on extremely large-scale text-image pair datasets, typically in the billion-level range. Their capability of producing high-fidelity images with straightforward text prompts facilitates their seamless adaptation to a range of generative tasks.

\begin{figure}[t]
   \centering
   \includegraphics[width=\linewidth]{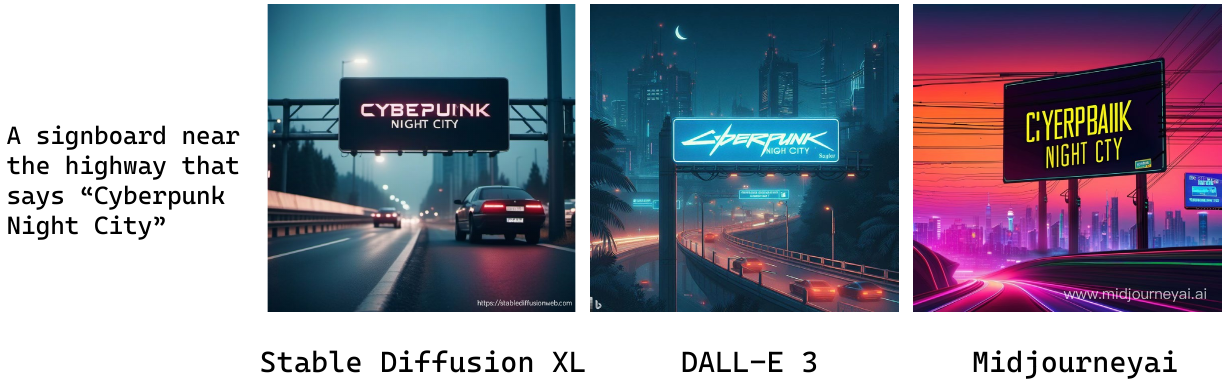}

   \caption{Text rendering problems of T2I models. The prompt we use is ``A signboard near the highway that says `Cyberpunk Night City'". Word spelling errors are commonly seen in images generated by Stable Diffusion XL, DALL-E 3 and Midjourneyai.}
   \vspace{-0.6cm}
   \label{fig:spelling}
\end{figure}

Although T2I generation models have made significant strides and can automate the process of artistic visual design to some extent, they still exhibit numerous limitations. For instance, when generating images that include human figures, these models often produce inaccurate or missing details in hands and faces. When synthesizing images with the desired text, these models often encounter serious spelling issues including incorrect, missing or repetitive characters, as illustrated in Fig. \ref{fig:spelling}. In some cases, they fail entirely to render text in generated images.
Some researchers~\cite{liu2022character} pointed out that these text rendering issues primarily stem from the inadequate information provided by the text encoder. They suggested that incorporating a character-aware text encoder with a large number of parameters (on the order of tens of billions) could mitigate this problem to some extent. The authors of DALL-E 3~\cite{Betker2023dalle3} also noted a limitation when the model encounters quoted text in a prompt: the T5 text encoder they utilize actually interprets tokens representing whole words and must map these to letters in an image, inevitably leading to unstable text rendering.

We suspect that those spelling issues in text synthesis is closely linked to the fundamental problems of existing T2I models, namely catastrophic neglect and incorrect attribute binding. To address this problem, we adopt and train a light-weight character-level text encoder to replace the original CLIP encoder employed in Stable Diffusion~\cite{rombach2022high}, thus providing more robust conditional guidance for the diffusion model. We then fine-tune a small portion of the model using the denoising score matching scheme and a proposed local attention map constraint. Finally, after implementing a refinement process during the inference stage, we shape the diffusion model into a powerful text designer capable of rendering precise words in images. Consequently, it can be utilized to precisely synthesize or edit text in arbitrary images based solely on text conditions.
We summarize our main contributions as follows:

\begin{itemize}
    \item We propose a diffusion model-based text image synthesis method, UDiffText, to address the text rendering challenges of existing T2I models. We leverages a character-level text encoder to derive robust text embeddings and employs a combination of the local attention loss and the scene text recognition loss to train our model on large-scale datasets. 
    \item The incorporation of segmentation map supervision offers a novel training strategy for T2I models, leading to enhanced text rendering performance. Experimental results demonstrate the effectiveness and superiority of our proposed method to the state of the art in terms of both text rendering accuracy and visual context coherency.
    \item As shown in Fig. \ref{fig:teaser}, we demonstrate several potential applications of our proposed UDiffText, including T2I generation with precise text content, arbitrary text generation as well as scene text editing.
\end{itemize}

\begin{figure*}[t]
  \centering
  \includegraphics[width=0.93\linewidth]{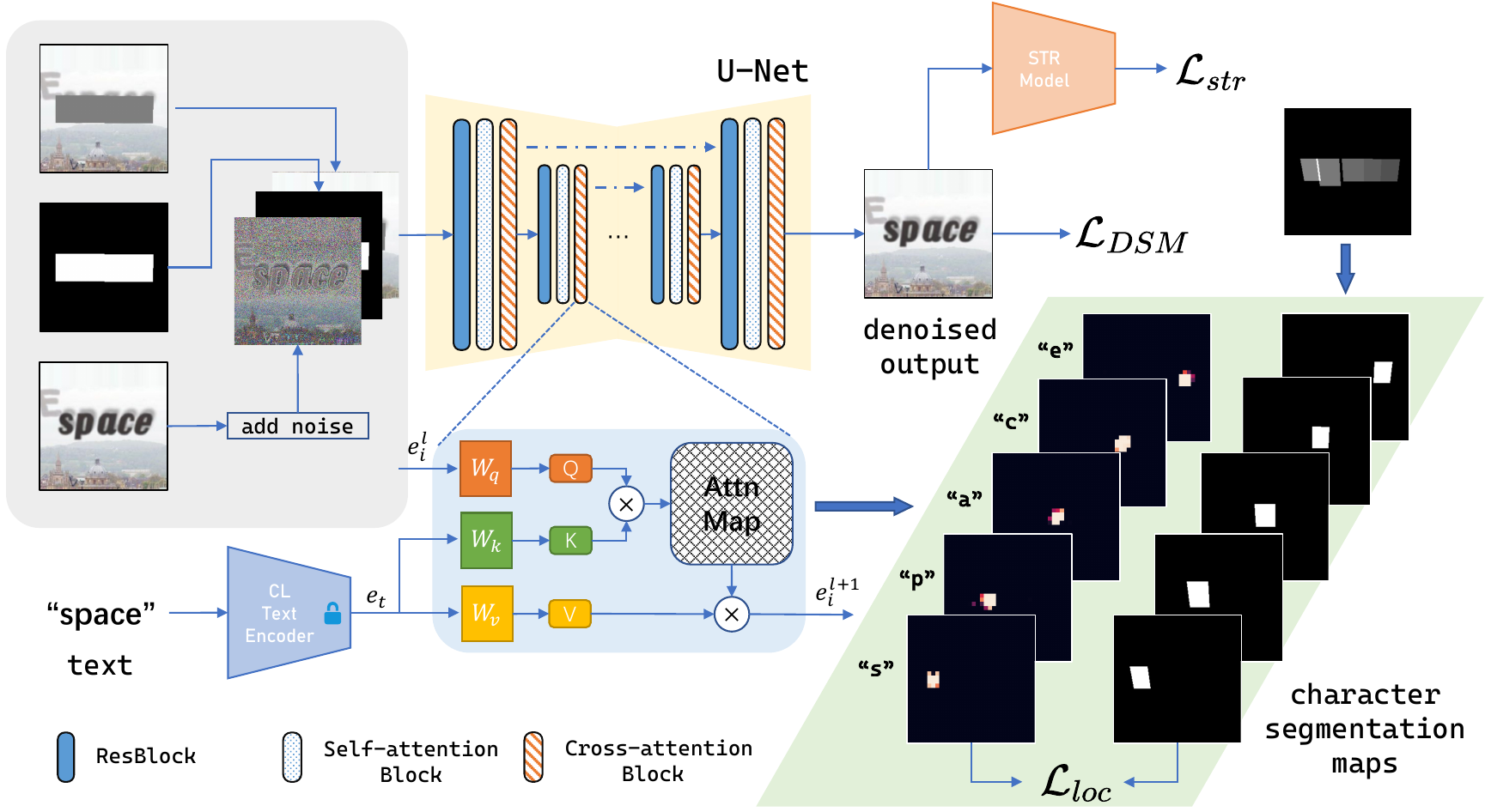}
  
  \caption{An overview of the training process of our proposed UDiffText. We build our model based on the inpainting version of Stable Diffusion (v2.0). A character-level (CL) text encoder is utilized to obtain robust embeddings from the text to be rendered. We train the model using denoising score matching (DSM) together with the local attention loss calculated based on character-level segmentation maps and the auxiliary scene text recognition loss. Note that only the parameters of cross-attention (CA) blocks are updated during training.}
  \vspace{-0.2cm}
  \label{fig:structure}
\end{figure*}
\section{Related Work}
\label{sec:related}

\subsection{Image Synthesis with Diffusion Models}

Recent state-of-the-art methods in image synthesis mostly utilize diffusion models (DMs). Ever since the introduction of denoising diffusion probability model (DDPM)~\cite{ho2020denoising}, large T2I models~\cite{ramesh2022hierarchical,saharia2022photorealistic,rombach2022high,podell2023sdxl,Betker2023dalle3} have achieved significant advancements in high-resolution image synthesis, exhibiting considerable diversity. Our research is conducted on the basis of Stable Diffusion~\cite{rombach2022high} and relevant efficient sampling algorithms~\cite{song2020score, karras2022elucidating}.

\subsection{Guided Diffusion}
\label{sec:guided}

While the advent of classifier-free guidance~\cite{ho2022classifier} has enhanced the generation performance of diffusion models, numerous methods have been explored to increase the controllability of these models using conditions from different modalities. Some approaches~\cite{nichol2021glide, saharia2022palette, brooks2023instructpix2pix} concatenate image conditions with noised latent variables as model input to furnish visual information. Others~\cite{gal2022image, kumari2023multi} utilize prompt tuning for concept-specific generation. Besides, certain methods~\cite{zhang2023adding, mou2023t2i} construct bypass network to control diffusion models using flexible pixel-domain conditions.

Notably, it is widely accepted that the cross-attention (CA) mechanism is pivotal in the generation process. Prompt-to-prompt~\cite{hertz2022prompt} evidences that CA maps are instrumental in determining the spatial layout of objects in generated images. Perfusion~\cite{tewel2023key} elaborates that the ``Keys'' in the CA mechanism govern the region of generated objects, while the ``Values'' dictate the features incorporated into the region. Structured Diffusion~\cite{feng2022training} employs noun phrase extraction to obtain more accurate CA features, thereby mitigating semantic attribute leakage. FastComposer~\cite{xiao2023fastcomposer} aligns CA maps with subject segmentation masks to address the problem of identity blending in multi-subject image generation. Attend-and-excite~\cite{chefer2023attend} directs diffusion models to refine the CA units to attend to all subject tokens in the text prompt, thus alleviating the issue of catastrophic neglect.
In this study, we attempt to constraint the CA maps of our diffusion model under the guidance of character-level segmentation maps to gain better text rendering performance.

\begin{figure}[t]
   \centering
   \includegraphics[width=\linewidth]{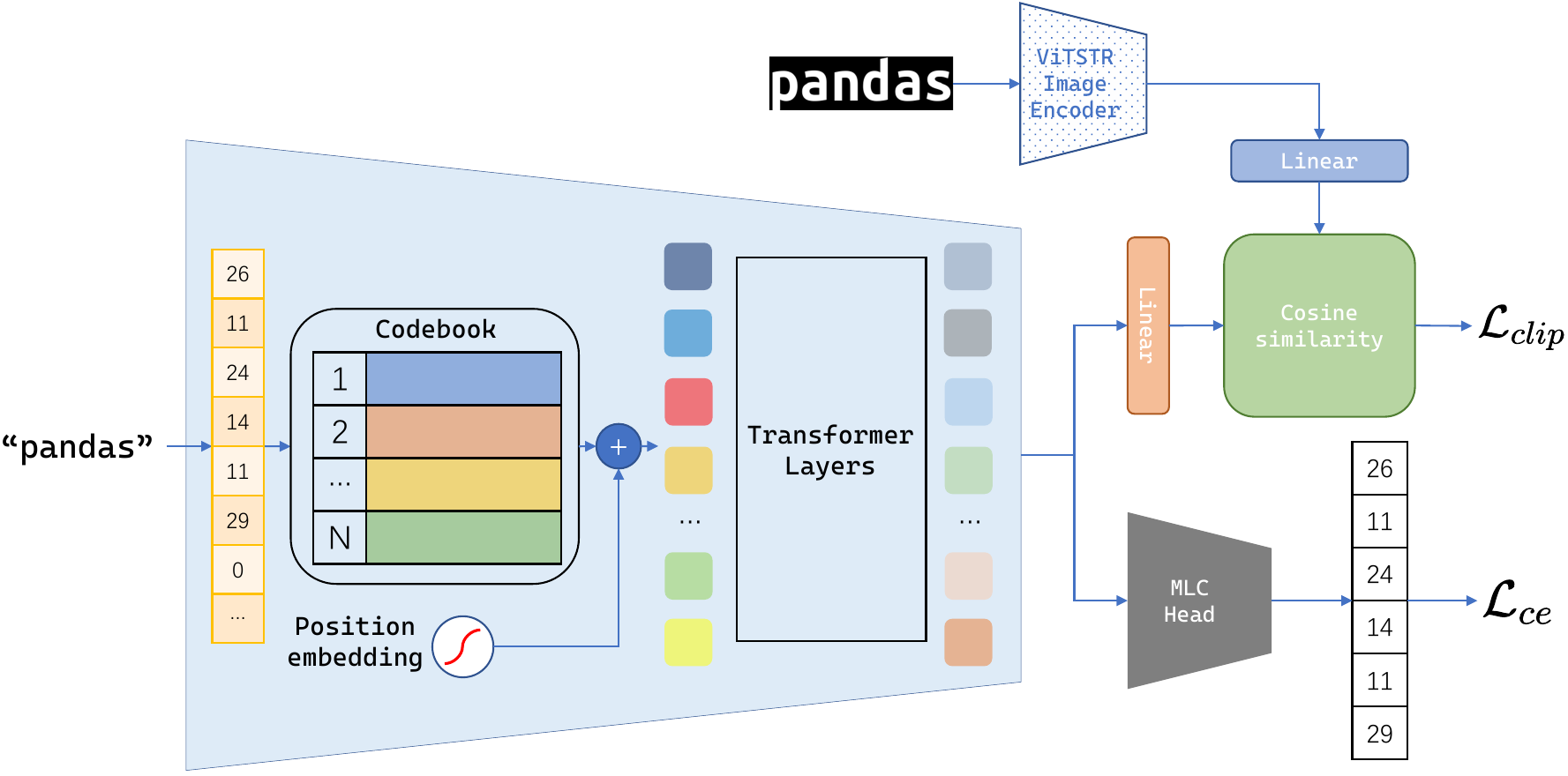}

   \caption{The network architecture of our character-level text encoder. A codebook is employed to translate the character indices into a sequence of learnable embeddings. These embeddings are enhanced by position embeddings and then input into a transformer to generate the encoded output.}
   \vspace{-0.6cm}
   \label{fig:encoder}
\end{figure}
\subsection{Scene Text Generation}

GAN-based scene text editing methods exhibit proficiency in generating coherent text within a specific visual context. STEFANN~\cite{roy2020stefann} constructs a FANnet to edit a single character and implements a placement algorithm to generate the expected word. SRNet~\cite{wu2019editing} and MOSTEL~\cite{qu2023exploring} divide the task into two primary parts: background inpainting and text style transfer. This division facilitates whole word editing in an end-to-end manner. Despite their simplicity and effectiveness, the capacity of these methods to generate high-resolution and polystylistic text images remains limited.

More recently, a number of approaches that aim to tackle the aforementioned text rendering challenges associated with diffusion models have been proposed. They leverage the robust capabilities of DMs to edit or generate scene text, thereby enhancing the quality and variety of the generated content.  DiffSTE~\cite{ji2023improving} uses the dual encoder structure (character text encoder and instruction text encoder) and performs instruction tuning to provide more accurate control for the backbone network. DiffUTE~\cite{chendiffute} uses an OCR-based glyph encoder to obtain glyph guidance from the rendered glyph image. Similarly, GlyphDraw~\cite{ma2023glyphdraw} leverages an additional image encoder and a fusion module to inject glyph condition and the fine-tuned model is able to generate images with coherent Chinese text. GlyphControl~\cite{yang2023glyphcontrol} applies ControlNet~\cite{zhang2023adding} to text image generation tasks by using the rendered reference image as both position and glyph guidance. TextDiffuser~\cite{chen2023textdiffuser} chooses to concatenate the segmentation mask as conditional input and uses the character-aware loss to control the generated characters more precisely.
In this study, we supplant the original CLIP text encoder in Stable Diffusion with a more robust character-level text encoder. This substitution equips the CA module with expressive and highly distinguishable character-aware embeddings. We firstly employ contrastive learning under visual supervision from a well-trained scene text recognition (STR) model to train the encoder. Then we fine-tune the CA blocks to yield more efficient CA ``Keys'' and ``Values'', which help the model generate more accurate text images.
\section{Method}
\label{sec:method}

As mentioned above, we aim to design a unified framework for high-quality text synthesis in both synthetic and real-world images. The proposed method, UDiffText, is built based on the inpainting variant of Stable Diffusion (v2.0). An overview of our method is depicted in Fig. \ref{fig:structure}. Specifically, we first design and train a light-weight character-level (CL) text encoder as a substitute for the original CLIP text encoder. Then, we train the model using the denoising score matching (DSM) loss in conjunction with the local attention loss and scene text recognition loss. More details of our proposed UDiffText will be elaborated in the following subsections.

\subsection{Character-level Text Encoder}
\label{sec:encoder}

As expounded in prior research~\cite{liu2022character}, a character-aware text encoder is deemed crucial in rectifying the issue of spelling errors in existing T2I models. However, the CLIP text encoder and T5 encoder, which are prevalently employed in T2I models, do not tokenize prompts at the character level. This results in the backbone network perceiving the entire word rather than its internal structure. A potential substitute for these encoders could be pre-trained character-aware transformers like ByT5~\cite{xue2022byt5}. However, only models with huge amounts of parameters (e.g., 20B) can exhibit reasonable performance, making the generation process inefficient and leading to unnecessary computational resource consumption. A possible solution is to utilize encoders to obtain character-level embeddings using pixel domain references. Yet, how to select appropriate references for the visual encoder is still an unsolved problem due to the requirement of a precise and generalized text representation to synthesize text images with diverse visual contexts. 

In this research, we design a CLIP-like text encoder that processes words at the character level. As shown in Fig. \ref{fig:encoder}, a target word is first mapped to corresponding indices and then converted into learnable embeddings using a codebook. Transformer layers are concatenated to produce the final output of shape $(B, L, d_{emb})$, where $B$ represents the batch size, $L$ indicates the maximum sequence length, and $d_{emb}$ denotes the dimension of the encoder. To obtain robust generalized embeddings, we train the text encoder $\mathcal{E}{text}$ using a combination of the contrastive loss and the multi-label classification loss. We first render the target word with a standard font style into an image. Then we use the ViTSTR~\cite{atienza2021vision} model, a scene text recognizer, as the image encoder $\mathcal{E}{image}$ to obtain robust visual features. A multi-label classification head $\mathcal{H}_{MLC}$ is trained concurrently to predict character indices $Ids$ from text embeddings. The calculation of the training loss is detailed in the following equations, where $\mathcal{T}$ and $\mathcal{I}_\mathcal{T}$ represent the text label and corresponding image, respectively, and $W_t, W_i$ are linear mapping matrices. We employ a cosine similarity ($\boldsymbol{CS}$) objective to align cross-modal features and use cross-entropy ($\boldsymbol{CE}$) as a multi-label classification loss to ensure that the learned embeddings are highly distinguishable:
\begin{gather}
    \mathbf{e}_{text} = \mathcal{E}_{text}(\mathcal{T}), \quad \mathbf{e}_{image} = \mathcal{E}_{image}(\mathcal{I}_\mathcal{T}), \label{eq:1}\\
    \mathcal{L}_{clip} = - \boldsymbol{CS}(W_t\mathbf{e}_{text}, W_i\mathbf{e}_{image}), \label{eq:2}\\
    \mathcal{L}_{ce} = \boldsymbol{CE}(\mathcal{H}_{MLC}(\mathbf{e}_{text}), Ids), \label{eq:3}\\
    \mathcal{L} = \mathcal{L}_{clip} + \lambda_{ce}\mathcal{L}_{ce}. \label{eq:4}
\end{gather}

\subsection{Training Strategy}
\label{sec:training}

Our system is constructed based on the inpainting version of Stable Diffusion~\cite{rombach2022high} (v2.0). During the training stage, the model functions as a denoiser, accepting a noised text image $\mathbf{x}_0 + \mathbf{n}$ of shape $(H, W)$, a binary mask $\mathcal{M}$ of the text region and the masked image $\mathbf{x}_\mathcal{M} = (\boldsymbol{J}-\mathcal{M})\odot\mathbf{x}_0$ as input ($\boldsymbol{J}$ is the all-ones matrix), and predicting the original text image as output. We utilize the denoising score matching (DSM) loss to train a denoiser for the specific text rendering task with the text condition $\mathcal{T}$:
\begin{equation}
    \mathcal{L}_{DSM} =\lambda_{\sigma}\left\|D_{\boldsymbol{\theta}}\left(\mathbf{x}_{0}+\mathbf{n} ; \sigma, \mathcal{T}, \mathcal{M}, \mathbf{x}_\mathcal{M} \right)-\mathbf{x}_{0}\right\|_{2}^{2},
\end{equation}
\noindent where $D_{\boldsymbol{\theta}}$ is a U-Net denoiser with the learnable parameter $\boldsymbol{\theta}$. $\left(\mathbf{x}_{0}, \mathcal{T}, \mathcal{M}\right) \sim p_{\text {data }}\left(\mathbf{x}_{0}, \mathcal{T}, \mathcal{M}\right)$ means that the text image, text label and binary mask of text region are randomly sampled from our dataset. $\mathbf{n} \sim \mathcal{N}\left(\mathbf{0}, \sigma^{2} \boldsymbol{I}_{d}\right)$ is the gaussian noise added to the text image and $\sigma$ represents the noise level. We set $\lambda_\sigma = \sigma^{-2}$ as the weighting function.

Our experimental results indicate that the DSM loss alone is insufficient to empower the model to render accurate text in generated images. This is mainly due to the fact that the $L2$ distance merely measures the mean distance between pixels, rather than the accuracy of character representation. To address this challenge, we incorporate a local attention loss to regulate the cross-attention maps of the model, a strategy 
similar to the approach adopted in~\cite{xiao2023fastcomposer}.

As mentioned in Sec. \ref{sec:guided}, we anticipate the model to acquire appropriate ``Keys'' and ``Values'' in cross-attention blocks. This enables the computed attention map to attend to corresponding character regions, and the learned character features could be appended to these regions. To achieve this goal, we utilize the supervision from character segmentation maps in our dataset. Specifically, for a character sequence $\mathcal{T}=\left\{\mathbf{c}^1, \mathbf{c}^2, \dots \mathbf{c}^L\right\}$, its corresponding segmentation map can be denoted as $\mathcal{S}_T = \left\{\mathbf{S}^1, \mathbf{S}^2, \dots \mathbf{S}^L\right\}$, where $\mathbf{S}^i$ of shape $(H, W)$ is a binary mask of the corresponding character $\mathbf{c}^i$ in the image. We can derive the attention maps $\mathcal{A}_i$ from each cross-attention block $i$ of the U-Net:
\begin{gather}
    \setlength{\abovedisplayskip}{3pt}
    \setlength{\belowdisplayskip}{3pt}
    \mathcal{Q}_i = W^Q_i \mathbf{e}_{image}, \text{ }\mathcal{K}_i = W^K_i \mathbf{e}_{text},  \text{ }\mathcal{V}_i = W^V_i \mathbf{e}_{text}, \\
    \mathcal{A}_i = \boldsymbol{softmax}\left(\mathcal{Q}_i\mathcal{K}_i^T /\sqrt{d}\right)\mathcal{V}_i.
\end{gather}
In this step, we partition the attention maps $\mathcal{A}_i$ on the dimension of sequence length into $\mathcal{A}_i = \left\{\mathbf{A}_i^1, \mathbf{A}_i^2, \dots \mathbf{A}_i^L\right\}$. Each $\mathbf{A}_i^j$ of shape $(H, W)$ represents the region of interest (ROI) of block $\mathbf{b}_i$ on the character $\mathbf{c}^j$. Subsequently, the local attention loss can be computed as follows:
\begin{equation}
    \setlength{\abovedisplayskip}{3pt}
    \setlength{\belowdisplayskip}{3pt}
    \begin{split}
        \mathcal{L}_{loc} = \frac1C \sum_{i=1}^{C} \left\{ \frac1L\sum_{j=1}^L\left(\boldsymbol{max} \left(\mathbb{G}\left(\mathbf{A}_{i}^{j}\right) \odot \left(\boldsymbol{J} - \mathbf{S}^{j}\right)\right)\right) \right.\\
        \left. - \frac1L\sum_{j=1}^L\left(\boldsymbol{max} \left(\mathbb{G}\left(\mathbf{A}_{ i}^{j}\right) \odot \mathbf{S}^{j}\right)\right) \right\},
    \end{split} 
\end{equation}
\noindent where $C$ represents the number of cross-attention blocks in the U-Net, $\mathbb{G}$ denotes a Gaussian blur and $\odot$ means the Hadamard product. The Gaussian blur is employed to perform low-pass filtering on the attention map, which helps to prevent excessive variance in the attended region. This approach ensures that the attention is distributed more evenly across the relevant regions, contributing to more accurate and stable model performance.

To enhance the text rendering accuracy, we incorporate the scene text recognition (STR) loss. Specifically, we employ a pre-trained STR model~\cite{bautista2022scene} to operate on the text region in the denoised results, and apply cross-entropy ($\boldsymbol{CE}$) to measure the correctness of the rendered word:
\begin{equation}
    \setlength{\abovedisplayskip}{3pt}
    \setlength{\belowdisplayskip}{3pt}
    \mathcal{L}_{str} =\boldsymbol{CE}\left(\boldsymbol{S}\left(D_{\boldsymbol{\theta}}\left(\mathbf{x}_{0}+\mathbf{n} ; \sigma, \mathcal{T}, \mathcal{M}, \mathbf{x}_\mathcal{M} \right) \odot \mathcal{M}\right), \mathcal{T}\right),
\end{equation}
\noindent where $\boldsymbol{S}$ represents the STR function, which accepts an RGB image as input and outputs the recognition logits.

During the training process, the majority of the U-Net parameters are frozen to maintain the fundamental image generation capability of the original model conditioned by the visual context. Only the parameters of the cross-attention map are updated to learn a generalized visual representation of each character in the character set. We refer to this type of model fine-tuning as ``knowledge complement''. In this fine-tuning stage, the model attends to the character regions of the text images and encodes the character shape and appearance into ``Keys'' and ``Values'' of the cross-attention map. The complete objective of our training strategy can be expressed as a combination of the denoising score matching (DSM) loss, the local attention loss and the scene text recognition loss:
\begin{equation}
    \setlength{\abovedisplayskip}{3pt}
    \setlength{\belowdisplayskip}{3pt}
    \mathcal{L} = \mathcal{L}_{DSM} + \lambda_{loc}\mathcal{L}_{loc} + \lambda_{str}\mathcal{L}_{str}.
\end{equation}

\vspace{-0.2cm}
\subsection{Refinement of Noised Latent}
\label{sec:refine}
Despite being constrained by the local attention loss, the fine-tuned model is still prone to producing spelling errors when rendering words in text images, such as missing some characters in a target word. We attribute this problem to a fundamental flaw in existing T2I models, i.e. catastrophic neglect. To address this issue, we implement noised latent refinement during the inference stage. Motivated by the generative semantic nursing approach introduced in~\cite{chefer2023attend}, we design a new loss function $\mathcal{L}_{aae}$ with the aim of maximizing the attention values of attention maps $\mathbf{A}_i^j$ corresponding to each character $\mathbf{c}^j$ within the region delineated by the binary mask $\mathcal{M}$:
\begin{equation}
    \setlength{\abovedisplayskip}{3pt}
    \setlength{\belowdisplayskip}{3pt}
    \begin{split}
        \mathcal{L}_{aae} (\mathcal{A}, \mathcal{M}) &= \\
        - \frac1C \sum_{i=1}^{C} &\left\{ \underset{1\le j \le N}{\boldsymbol{min}}\left(\boldsymbol{max} \left(\mathbb{G}\left(\mathbf{A}_{i}^{j}\right) \odot \mathcal{M}\right)\right) \right\}.
    \end{split}
\end{equation}

Our noised latent refinement process mainly consists of two steps: identifying an optimal initial noise $\mathbf{n}$, and optimizing the noised latent $\mathbf{z}_t$ at each timestep $t$. Initially, we sample Gaussian noise $N$ times from the distribution $\mathcal{N}\left(\mathbf{0}, \sigma^{2} \boldsymbol{I}{d}\right)$. For each sampled noise, we swiftly execute the entire denoising process in a limited number (e.g., 2) of iterations and compute the corresponding objective $\mathcal{L}_{aae}$ at the final timestep. Subsequently, we select the noise with the minimum loss value as our initial noise $\mathbf{n}_{i^*}$. During the denoising process to get the final output, we refine the noised latent $\mathbf{z}_t$ using the gradient calculated based on the proposed objective $\mathcal{L}_{aae}$ at each timestep $t$:
\begin{equation}
    \setlength{\abovedisplayskip}{3pt}
    \setlength{\belowdisplayskip}{3pt}
    \mathbf{z}_t' = \mathbf{z}_t - \alpha_t \cdot\nabla_{\mathbf{z}_t}\mathcal{L}_{aae},
\end{equation}
\noindent where $\alpha_t$ represents the learning rate used to update the noised latent $\mathbf{z}_t$ at each timestep $t$. The gradient $\nabla_{\mathbf{z}_t}\mathcal{L}_{aae}$ is computed in a backward manner through the parameters of the U-Net on the noised latent $\mathbf{z}_t$. The specifics of the refinement process are outlined in Alg. \ref{alg:refinement}. We utilize the denoising algorithm proposed in~\cite{karras2022elucidating}. Here, $\boldsymbol{EulerStep}$ denotes a single sampling step implemented using the Euler's method and $\boldsymbol{ODESchedule}(T)$ is the ODE scheduler which takes the number of ODE solver iterations $T$ as input and outputs the $\sigma$s of discretized sampling steps.

\begin{algorithm}[htb]
    \setstretch{1.0}
    \setlength{\textfloatsep}{0.1cm}
    \setlength{\floatsep}{0.1cm}
	\renewcommand{\algorithmicrequire}{\textbf{Input:}}
	\renewcommand{\algorithmicensure}{\textbf{Output:}}
	\renewcommand{\algorithmiccomment}[1]{\hfill $\triangleright$ #1}
	\caption{Denoising process with refinement}
	\label{alg:refinement}
	\begin{algorithmic}[1]
		\REQUIRE A binary mask $\mathcal{M}$, a text condition $\mathcal{T}$, a masked image $\mathbf{x}_\mathcal{M} = (\boldsymbol{J}-\mathcal{M})\odot\mathbf{x}_0$, a U-Net denoiser $D_{\boldsymbol{\theta}}$ and a latent decoder $\mathcal{D}$
        \ENSURE the denoised image $\hat{\mathbf{x}_0}$
        
        \STATE $\{\sigma_2, \sigma_1\} \leftarrow \boldsymbol{ODESchedule}(2)$
        
        \FOR {$i \leftarrow 1,2\dots N$} 
        \STATE $\mathbf{n}_i \sim \mathcal{N}\left(\mathbf{0}, \sigma_2^{2} \boldsymbol{I}_{d}\right)$  
        \STATE $\mathbf{d}, \mathcal{A}_2 \leftarrow D_{\boldsymbol{\theta}}(\mathbf{n}_i; \sigma_2, \mathcal{T}, \mathcal{M}, \mathbf{x}_\mathcal{M})$
        \STATE $\mathbf{z} \leftarrow \boldsymbol{EulerStep}(\mathbf{d}, \mathbf{n}_i, \sigma_2)$
        \STATE $\_, \mathcal{A}_1 \leftarrow D_{\boldsymbol{\theta}}(\mathbf{z}; \sigma_1, \mathcal{T}, \mathcal{M}, \mathbf{x}_\mathcal{M})$
        \STATE $\mathcal{L}_i \leftarrow \mathcal{L}_{aae}(\mathcal{A}_1, \mathcal{M})$ 
        \ENDFOR
        
        \STATE $i^* \leftarrow \underset{1\le i \le N}{argmin} \text{ }\mathcal{L}_i$
        \COMMENT{select the best initial noise}
        \STATE $\mathbf{z}_T \leftarrow \mathbf{n}_{i^*}$ 
        \STATE $\{\sigma_T, \sigma_{T-1}, \dots \sigma_1\} \leftarrow \boldsymbol{ODESchedule}(T)$
        
        \FOR {$t \leftarrow T, T-1\dots 1$}
        \STATE $\_, \mathcal{A}_t \leftarrow D_{\boldsymbol{\theta}}(\mathbf{z}_t; \sigma_t, \mathcal{T}, \mathcal{M}, \mathbf{x}_\mathcal{M})$
        \STATE $\mathcal{L}_t \leftarrow \mathcal{L}_{aae}(\mathcal{A}_t, \mathcal{M})$ 
        \STATE $\mathbf{z}_t' \leftarrow \mathbf{z}_t - \alpha_t \cdot \nabla_{\mathbf{z}_t}\mathcal{L}_t$
        \COMMENT{refine the noised latent}
        \STATE $\mathbf{d}_{t-1}, \_ \leftarrow D_{\boldsymbol{\theta}}(\mathbf{z}_t'; \sigma_t, \mathcal{T}, \mathcal{M}, \mathbf{x}_\mathcal{M})$
        \STATE $\mathbf{z}_{t-1} \leftarrow \boldsymbol{EulerStep}(\mathbf{d}_{t-1}, \mathbf{z}_t', \sigma_t)$
        \ENDFOR
        
        \STATE $\hat{\mathbf{x}_0} \leftarrow \mathcal{D}(\mathbf{z}_0)$
        
        \RETURN $\hat{\mathbf{x}_0}$
	\end{algorithmic}
\end{algorithm}
\section{Experiments}
\label{sec:experiments}

\subsection{Datasets and Evaluation Metrics}

To apply the training strategy mentioned in Sec. \ref{sec:training} and enhance the generalization capability of our proposed model, we require large-scale datasets, which should offer a diverse range of character samples, varying in shape and style. Ideally, the datasets should contain large numbers of text images, text annotations and the bounding boxes of text regions. Additionally, the character-level segmentation maps are also necessary. Considering these requirements, we have selected two datasets to constitute our training data:

\begin{itemize}
    \item \textbf{SynthText in the Wild}~\cite{Gupta16} is a synthetically generated dataset, in which word instances are placed in natural scene images, while taking into account the scene layout. The dataset consists of 800,000 images with approximately 8 million synthetic word instances. Each text instance is annotated with its text-string, word-level and character-level bounding-boxes, which we utilize to generate character-level segmentation maps.
    
    \item \textbf{LAION-OCR}~\cite{chen2023textdiffuser} derives from the large-scale dataset LAION-400M~\cite{schuhmann2021laion}. It contains 9,194,613 filtered high-quality text images including advertisements, notes, posters, covers, memes, logos, etc. The authors of~\cite{chen2023textdiffuser} trained a character-level segmentation model to obtain the segmentation maps of the text images.
\end{itemize}

For the purpose of validation, we gather datasets that include text images not previously encountered by the model. These datasets are derived from various tasks, encompassing scene text detection and segmentation.

\begin{itemize}
    \item \textbf{ICDAR13}~\cite{karatzas2013icdar} is the standard benchmark for evaluating near-horizontal text detection, which contains 233 test images.
    
    \item \textbf{TextSeg}~\cite{xu2021rethinking} is a multi-purpose text dataset focused on segmentation. It contains real-world text images collected from posters, greeting cards, covers, logos, road signs, billboards, digital designs, handwriting, etc. 340 images of them are for validation.
    
    \item \textbf{LAION-OCR} evaluation dataset. We partition a subset of the \textbf{LAION-OCR} dataset for the purpose of validation. The text images in this subset are not exposed to the model during the training phase.
\end{itemize}

\begin{table*}[htb]
    \centering
    \resizebox{\linewidth}{!}{
    \begin{tabular}{@{}l|cccc|cccc|c|c@{}}
     \toprule[1.5pt]
     \multirow{2}{*}{\textbf{Method}} & \multicolumn{4}{c|}{\textbf{SeqAcc-Recon (\%)}$\uparrow$} & \multicolumn{4}{c|}{\textbf{SeqAcc-Editing (\%)}$\uparrow$} & \multirow{2}{*}{\textbf{FID}$\downarrow$} & \multirow{2}{*}{\textbf{LPIPS}$\downarrow$}\\ 
     \cmidrule(lr){2-5} \cmidrule(lr){6-9}
     & \textbf{ICDAR13 (8ch)} & \textbf{ICDAR13} & \textbf{TextSeg} & \textbf{LAION-OCR} & \textbf{ICDAR13 (8ch)} & \textbf{ICDAR13} & \textbf{TextSeg} & \textbf{LAION-OCR} & & \\
     \midrule[1pt]
     MOSTEL & 75.0& 68.0& 64.0& 71.0& 35.0& 28.0& 25.0& 44.0& 25.09& 0.0605\\
     SD-Inpainting & 32.0& 29.0& 11.0& 15.0& 8.0& 7.0& 4.0& 5.0& 26.78& 0.0696\\
     DiffSTE & 45.0& 37.0& 50.0& 41.0& 34.0& 29.0& 47.0& 27.0& 51.67& 0.1050\\
     TextDiffuser & 87.0& 81.0& 68.0& 80.0& 82.0& 75.0& 66.0& 64.0& 32.25& 0.0834\\
     \midrule[1pt]
     Ours & \textbf{94.0}& \textbf{91.0}& \textbf{93.0}& \textbf{90.0}& \textbf{84.0}& \textbf{83.0}& \textbf{84.0}& \textbf{78.0}& \textbf{15.79}& \textbf{0.0564}\\
     \bottomrule[1.5pt]
    \end{tabular}}
    \caption{Quantitative comparison between our method and four baselines. \textbf{ICDAR13 (8ch)} denotes that we restrict the text length to no more than 8 characters for the purpose of evaluating short word rendering performance. The best scores are highlighted in bold.}
    \vspace{-0.4cm}
    \label{tab:quan results}

\end{table*}

We assess the performance of our proposed model in two aspects: image quality and text sequence accuracy. For the evaluation of image quality, we employ Fréchet Inception Distance (\textbf{FID})~\cite{heusel2017gans} to measure the distance between the text images in the dataset and the images generated by our model and other baselines.  Furthermore, we incorporate Learned Perceptual Image Patch Similarity (\textbf{LPIPS})~\cite{zhang2018unreasonable} as an additional metric to assess the quality of the generated images. The above metrics provide an indication of the visual coherence between the rendered text and its background. Given that our primary objective is to correct word spelling errors prevalent in existing diffusion models, we utilize an off-the-shelf scene text recognition (STR) model~\cite{bautista2022scene} to identify the rendered text. Subsequently, we employ sequence accuracy (\textbf{SeqAcc}) to evaluate the word-level correctness by comparing the STR result with the ground truth.

\subsection{Implementation Details}
UDiffText primarily comprises two components: a U-Net backbone and the proposed character-level text encoder.
For the U-Net, we employ the pre-trained checkpoint of Stable Diffusion (v2.0) inpainting version. The model is fine-tuned using an image size of $512\times 512$ on the SynthText dataset for 100k steps and then on the LAION-OCR dataset for an additional 100k steps. The training process utilizes a batch size of 64 and a learning rate of $5\times 10^{-5}$. The U-Net encompasses 891M parameters, of which only 75.9M (the parameters of the cross-attention layers) are updated during training.
As for the character-level text encoder, it undergoes initial training using the strategy outlined in Sec. \ref{sec:encoder} for 8k steps with a batchsize of 256 and a learning rate of $1\times 10^{-5}$. Following this, it is frozen and connected to the U-Net for subsequent training. The proposed encoder comprises approximately 302M parameters.
In the training stage, we set $\lambda_{ce}$ to 0.1, $\lambda_{loc}$ to 0.01 and $\lambda_{str}$ to 0.001. During the inference stage, we employ 50 sampling steps and utilize a classifier-free guidance (CFG) scale of 5.0.

\begin{figure}[htb]
   \centering
   \includegraphics[width=\linewidth]{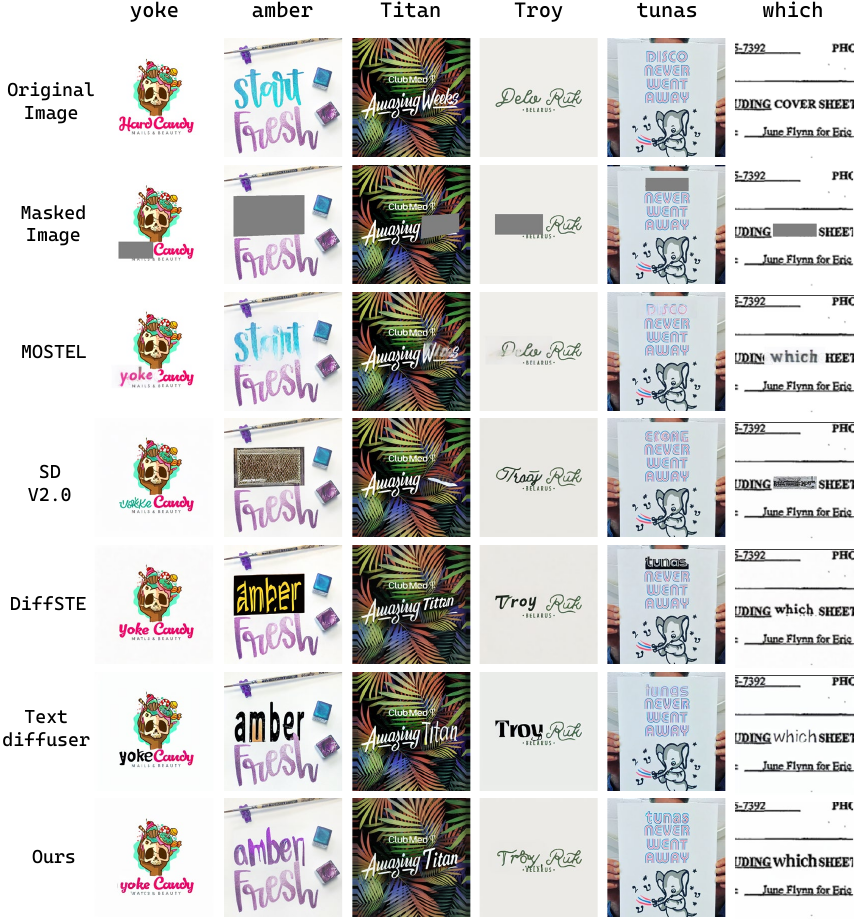}

   \caption{Qualitative results on the scene/document/poster text editing task. The first row consists of the original images, while the second row comprises the input images with binary masks applied to the text region. The specific word to be generated is indicated at the top of each column.}
   \vspace{-0.6cm}
   \label{fig:results}
\end{figure}

\subsection{Quantitative and Qualitative Results}
To validate the superiority of our proposed method, we compare it with several scene text generation/editing techniques, including the GAN-based method (MOSTEL~\cite{qu2023exploring}), and diffusion-based methods (DiffSTE~\cite{ji2023improving} and TextDiffuser~\cite{chen2023textdiffuser}). For better comparison, we evaluate all methods across two distinct tasks: scene text reconstruction and scene text editing. In the case of the former, we employ the models to reconstruct the text image using the provided ground truth text label and binary mask. For the latter, we substitute the original text in each image with a random word of equivalent length and evaluate the models by generating images containing the edited text. The sequence accuracy ($\textbf{SeqAcc}$) for these tasks is denoted as \textbf{SeqAcc-Recon} and \textbf{SeqAcc-Editing}, respectively. We limit the text length in each instance to a maximum of 12 characters and randomly select 100 images from each dataset for testing.

The quantitative comparison results are presented in Tab. \ref{tab:quan results}. For TextDiffuser~\cite{chen2023textdiffuser}, we utilize their inpainting variant, where we render the desired text in a standard font (Arial) at the masked region as the input for their proposed segmentor. As for MOSTEL~\cite{qu2023exploring}, we employ it to generate the text at the masked region and then integrate the output back into the original image. Their \textbf{FID} and \textbf{LPIPS} scores appear satisfactory, in part because the background remains unaltered. Furthermore, we also assess the performance of the pre-trained Stable Diffusion (v2.0) inpainting version as a baseline result. We set the prompt as ``[word to be rendered]'' for fair comparison. Overall, our method outperforms the baselines across all quantitative metrics, suggesting that our proposed model is capable of generating text images with superior sequence accuracy and quality, conditioned solely on the text label. For the qualitative results, we display the outputs of all aforementioned methods on the scene text editing task. As illustrated in Fig. \ref{fig:results}, our method yields the most visually pleasing results, characterized by high text rendering accuracy and visual context coherency For more qualitative results, please refer to Sec. \ref{more compare} of our supplementary material.

\subsection{Ablation Study}

\begin{table}
  \centering
  \resizebox{0.6\linewidth}{!}{\begin{tabular}{@{}lc@{}}
    \toprule
    \textbf{Setting} & \textbf{SeqAcc-Recon (\%)}$\uparrow$\\
    \midrule
    Base & 8.0\\
    + CL encoder & 40.0\\
    + $L_{loc}$ & 54.0\\
    + $L_{str}$ & 65.0\\
    + Refinement & \textbf{76.0}\\
    \bottomrule
  \end{tabular}}
  \caption{Ablation study results on different settings. In each case, we utilize the model to reconstruct the text in synthetic images and evaluate the performance using the sequence accuracy metric.}
  \vspace{-0.6cm}
  \label{tab:ablation}
\end{table}

\begin{figure}[t]
   \centering
   \includegraphics[width=\linewidth]{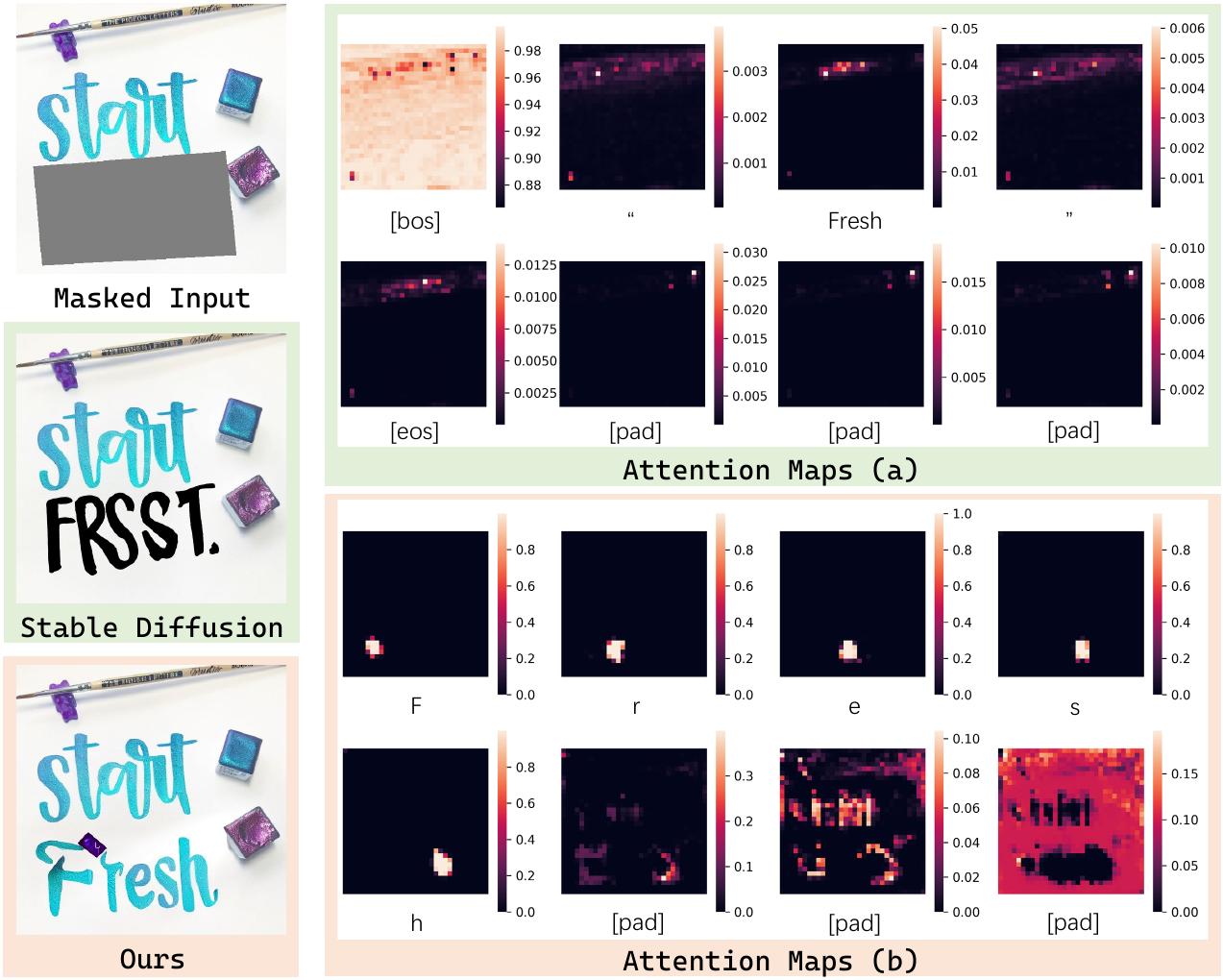}

   \caption{Visualization results. The expected text is ``Fresh” and the masked input is displayed at the top left. The attention maps extracted from the U-Net of Stable Diffusion (a) and ours (b) can be observed on the right side. The specific token of each attention map is annotated at the bottom.}
   \vspace{-0.6cm}
   \label{fig:ablation}
\end{figure}

To assess the efficacy of each design choice in our method, we perform an ablation study on various settings, which include:
(1) Base: The inpainting version of the pre-trained Stable Diffusion (v2.0), which uses the CLIP text encoder to obtain conditional embeddings.
(2) CL Encoder: We employ our proposed character-level text encoder (CL Encoder) as a replacement for the CLIP encoder.
(3) $L_{loc}$: We incorporate the proposed local attention loss into the basic diffusion loss to serve as the training objective.
(4) $L_{str}$: We introduce the scene text recognition loss for additional supervision.
(5) Refinement: We apply the refinement of noised latent, as mentioned in Sec. \ref{sec:refine}, at the inference stage to enhance text accuracy.
We train the model under all the above settings on the SynthText dataset for 6k steps and test them on the corresponding evaluation set. The quantitative results of sequence accuracy (\textbf{SeqAcc}) are presented in Tab. \ref{tab:ablation}, which indicate that the whole model outperforms the other variants.

To further illustrate the efficacy of our proposed character-level text encoder and local attention loss, we compare the performance of our UDiffText with that of Stable Diffusion. In a specific generation scenario, we extract the attention maps from the U-Net model during an intermediate inference step. As depicted in Fig. \ref{fig:ablation}, it is evident that our UDiffText focuses on the precise regions of each rendered character, whereas Stable Diffusion exhibits ambiguous attention areas within the rendered word, leading to incorrect results and attention maps devoid of meaningful information. This experiment indicates that the local attention loss indeed imposes an effective constraint on the attention maps, thereby enhancing the interpretability of our proposed method. More visualization analysis is available in Sec. \ref{more visual} of our supplementary material.

\subsection{Applications}
\textbf{Scene text generation/editing.}
Taking an arbitrary image, a binary mask and a text sequence as input, UDiffText generates a modified image with the desired text rendered in a specific region defined by the mask. This inpainting-based architecture makes the proposed method suitable for a variety of inpainting-like text rendering applications. As demonstrated in Fig. \ref{fig:teaser} (a)(b) and Fig. \ref{fig:results}, our method can be applied to tasks involving the generation or editing of scene text in real-world images and scanned documents. Obviously, the proposed UDiffText has the potential to be applied to construct large-scale scene text image datasets, given its capability to generate context-coherent text images that do not exist in the real world. Moreover, our UDiffText can also be applied to graphic design tasks like poster design and advertisement design.

\noindent \textbf{T2I generation with accurate text content.} Leveraging the text editing capability of our proposed model, we devise a two-stage method for T2I generation that ensures accurate text rendering, as shown in Fig. \ref{fig:teaser} (c). Specifically, in our experiments, we first utilize a pre-trained T2I model (\cite{podell2023sdxl} or \cite{Betker2023dalle3}) to produce a preliminary result using the prompt template generated by LLM. Then, we employ a scene text detector~\cite{ye2023deepsolo} to mask the text region in the generated image. At last, our UDiffText is applied to the masked image to produce the final output, which features accurate text and a consistent style (see Fig. \ref{fig:teaser} (c) and Sec. \ref{more app} of our supplemental material). Furthermore, we also quantitatively evaluate our method using the SimpleBench prompt templates proposed in~\cite{yang2023glyphcontrol}. Experimental results show that our approach significantly improve the average text rendering accuracy of the pre-trained SDXL model from 8.0\% to 60.0\%.
\section{Limitations}
\label{sec:limitation}

Since our model relies on visual context to render the expected text, it may struggle to generate coherent text when the image background is relatively simple. Furthermore, the current version of our method can only satisfactorily handle text sequences with a limited number of characters (up to 12 characters in our implementation). This limitation may affect the performance of our method in tasks that require longer text inputs, such as paragraph generation or long document editing (see some examples of failure cases in our supplementary material). One possible solution to address this problem is to synthesize the long text sequence word by word.
\section{Conclusion}
\label{sec:conclusion}

In this paper, we proposed UDiffText, a novel method for high-quality text synthesis in arbitrary images using character-aware diffusion models. We designed and trained a character-level text encoder that provides robust text embeddings and fine-tuned the diffusion model with local attention control and scene text recognition supervision. Our method can generate coherent images with accurate text and can be used for arbitrary text generation, scene text editing and T2I generation with precise text content. We demonstrated the effectiveness of our method through extensive experiments and comparisons with existing methods, showing the superiority of the proposed UDiffText to the state of the art in terms of both text rendering accuracy and visual context coherency. In the future, we plan to explore more ways to improve the controllability and diversity of our method, and extend it to other text-related image synthesis tasks.
{
    \small
    \bibliographystyle{ieeenat_fullname}

}

\clearpage
\setcounter{page}{1}
\maketitlesupplementary

\setcounter{section}{0}
\renewcommand\thesection{\Alph{section}}

\section{More Implementation Details}

We have observed that the proportion of the masked region in a given image significantly influences the performance of text rendering. Consequently, we strictly constrain the proportions of the text mask and character segmentation mask in our training datasets. In particular, images with a text mask proportion of less than 1\% or a character segmentation mask proportion of less than 0.1\% are filtered out. Additionally, we perform image cropping and resizing to achieve inputs of a uniform scale and to maintain the proportion of the text region within a reasonable range.
 
\begin{figure*}[ht]
   \centering
   \vspace{-0.8cm}
   \includegraphics[width=0.9\linewidth]{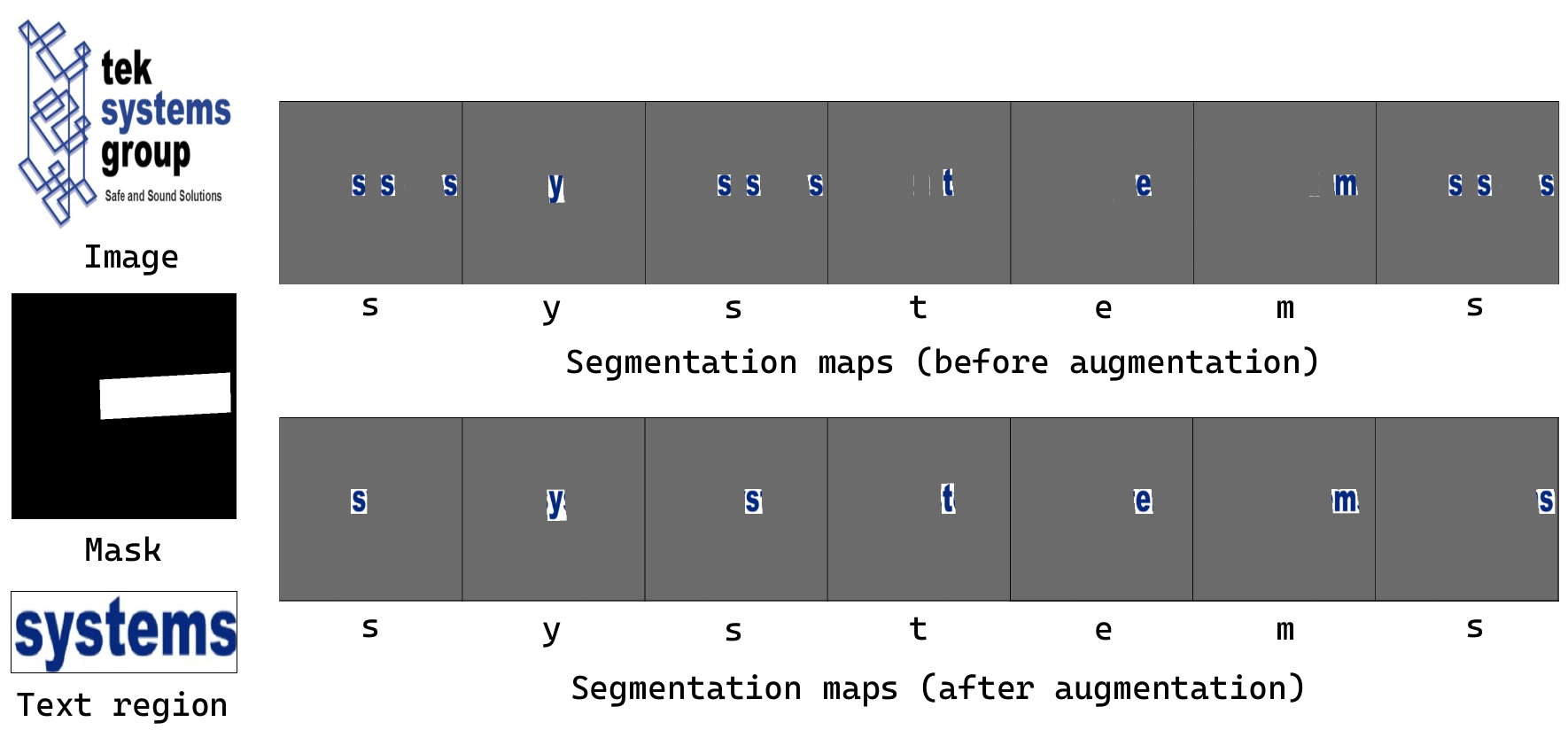}

   \caption{Data augmentation. The image and the binary mask of the text region can be seen at the left side while the corresponding segmentation map for each character is shown at the right side.}
   \label{fig:augment}
\end{figure*}

For images in LAION-OCR, character-level segmentation maps are derived using the segmentation model proposed in~\cite{chen2023textdiffuser}. This model is position-unaware and assigns the same index to all regions of a specific character (e.g., “a”). Besides, the segmentation model may produce unsatisfactory or incorrect results, such as omitting certain characters or partially masking them. To perform data cleaning and augmentation, we initially employ connected components extraction to separate repeated characters in the binary masks, thereby providing positional information and eliminating ambiguity in attention map constraints. Subsequently, we apply a morphological opening operation to eliminate noise points and use morphological dilation to slightly expand the masked character areas for improved supervision. An illustrative example of the data augmentation process can be seen in Fig. \ref{fig:augment}. It should be noted that the issue of missing characters in segmentation maps cannot be completely resolved and may adversely affect the text rendering performance of our model.

\section{More Comparison Results}
\label{more compare}
\begin{figure*}[htb]
   \centering
   \includegraphics[width=0.9\linewidth]{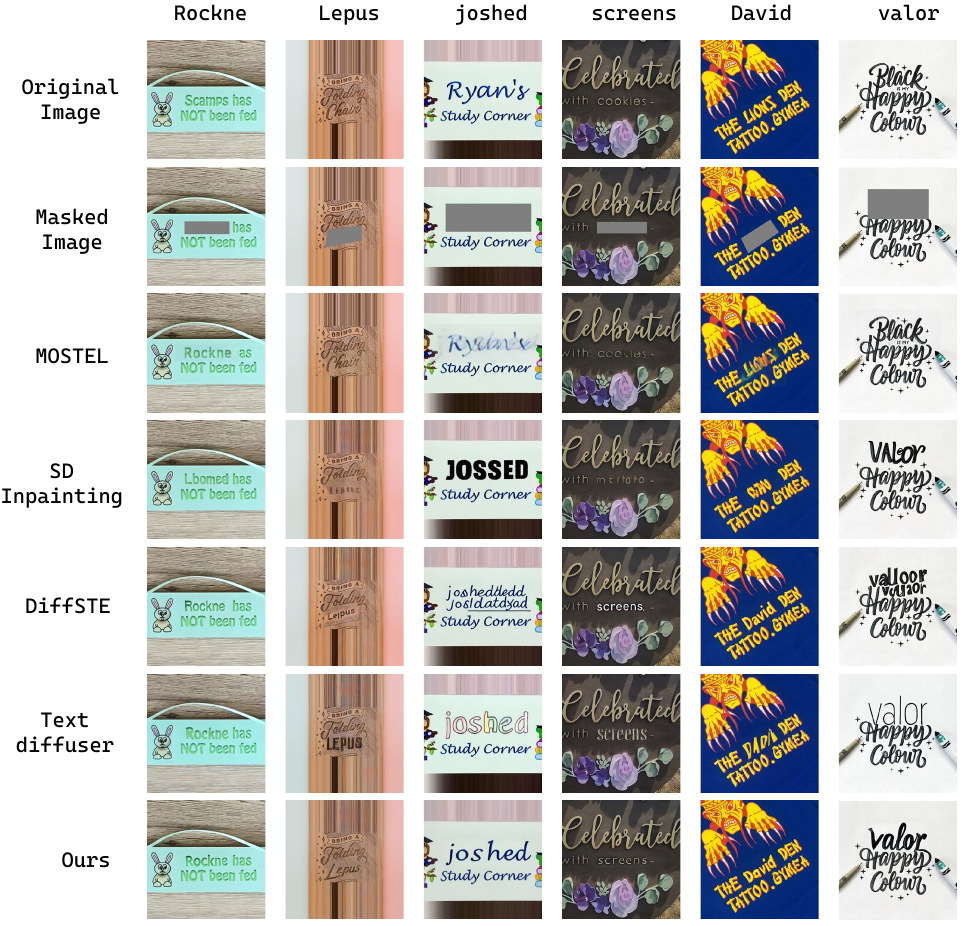}

   \caption{Additional comparison results on the scene text editing task. The first row consists of the original images, while the second row comprises the input images with binary masks applied to the text region. The specific word to be generated is indicated at the top of each column.}
   \label{fig:add_compare}
\end{figure*}

We carry out additional qualitative experiments on the scene text editing task and compare our results with those of the aforementioned baselines. Further results can be viewed in Fig. \ref{fig:add_compare}. It is evident that our method produces the most visually appealing outcomes, distinguished by high text rendering precision and consistency with the visual context.

\section{More Application Results}
\label{more app}
\begin{figure*}[htb]
   \centering
   \includegraphics[width=0.9\linewidth]{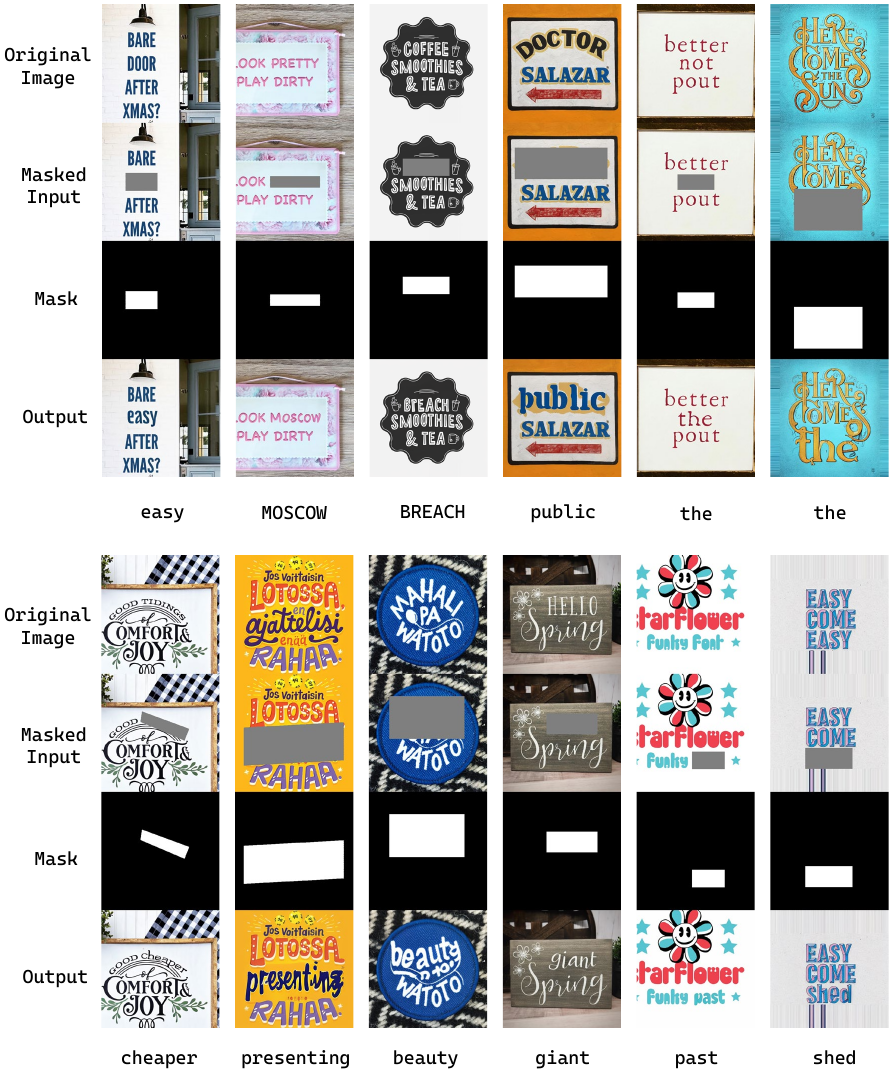}

   \caption{Additional application results for scene text editing task. The word to be rendered is annotated at the bottom of each case.}
   \label{fig:add_app1}
\end{figure*}

\begin{figure*}[htb]
   \centering
   \includegraphics[width=0.85\linewidth]{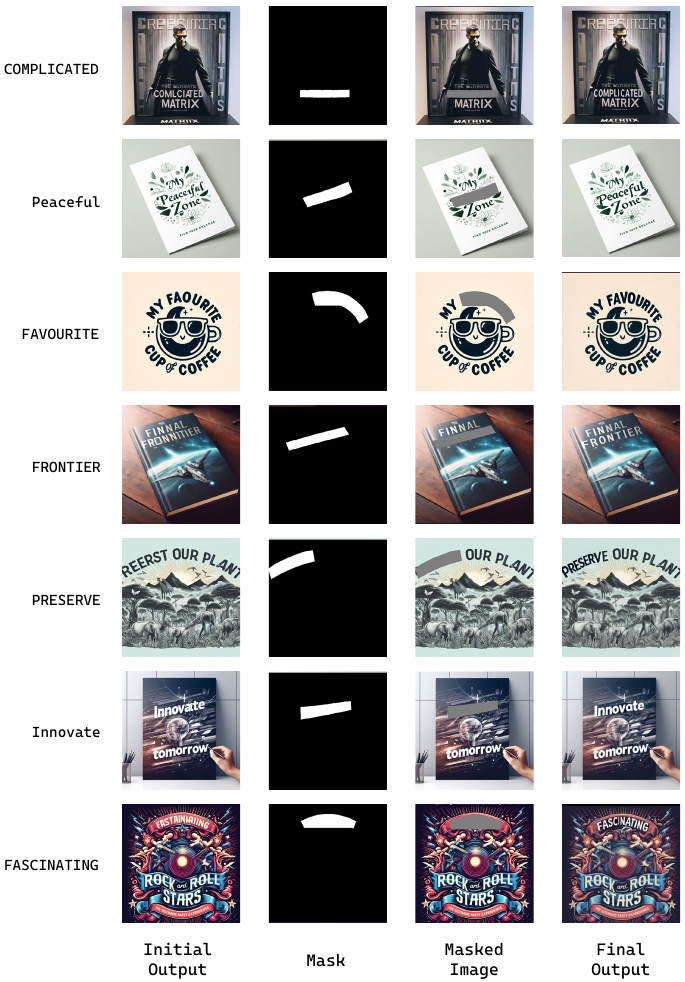}

   \caption{Additional application results for accurate T2I generation task. The first column demonstrates the initial outputs of DALL-E-3 conditioned by the given prompts while the last column shows our final outputs after correcting the text in masked regions. The word to be corrected is annotated at the left of each case.}
   \label{fig:add_app2}
\end{figure*}

We present additional qualitative results on the previously discussed scene text editing task (Fig. \ref{fig:add_app1}) and the accurate T2I generation task (Fig. \ref{fig:add_app2}). Leveraging the inpainting-based architecture, UDiffText is proficient in generating coherent text in both real-world images and AI-generated images. Consequently, it can serve as an artistic text designer in a variety of graphic design tasks, including poster design and advertisement design.

In the accurate T2I generation task, we utilize off-the-shelf LLM (GPT-3.5) to generate the prompts for first-stage image generation, the prompt we use for each case are as follows:

\begin{enumerate}
    \item A poster for a movie premiere with the title ``Complicated Matrix'' and the tagline ``The ultimate choice is yours''. The poster has an image of a man in a black suit and sunglasses holding a gun. The text is in a futuristic and metallic font.
    \item A flyer for a yoga class with the title ``My Peaceful Zone'' and the slogan ``Find your balance''. The flyer has a white background with green leaves and flowers. The text is in a simple and elegant font.
    \item A logo for a coffee shop called ``My Favourite cup of coffee''. The logo is a stylized coffee bean with a smiley face and sunglasses. The text is in a handwritten and casual font.
    \item A book cover for a sci-fi novel called ``The Final Frontier''. The book cover has an image of a spaceship flying over an alien planet. The text is in a futuristic and metallic font.
    \item Create an artistic composition for a nature conservation campaign. Include lush landscapes, endangered species, and the phrase ``Preserve Our Planet'' in elegant typography.
    \item Craft a captivating banner for a technology summit. Use sleek lines, futuristic elements, and include the phrase ``Innovate for Tomorrow'' in a dynamic font.
    \item A poster for a music festival with the title ``Fascinating rock and roll stars'' and the logo of a guitar. The poster has a colorful background with geometric shapes and patterns. The text is in a bold and funky font.
\end{enumerate}

\section{More Visualization Results}
\label{more visual}
To provide a more intuitive demonstration of the proposed local attention constraint, we present additional visualization results in Fig. \ref{fig:add_visual}. For a specific text rendering case, we extract the attention maps from the middle block of the U-Net at an intermediate sampling step. It is clear that under the constraint of our local attention loss, the model focuses on the specific region of each character. The attention values are high and centralized in the character areas, while they are nearly zero in areas of no concern. This type of constraint assists the model in concentrating on learning the visual features of characters rather than irrelevant textures. Furthermore, we up-sample the attention maps to the scale of the output image and obtain segmentation maps of each generated character, as shown in the last column. This experiment illustrates a potential application of text segmentation based on our trained model and corresponding image editing methods with diffusion models.

\begin{figure*}[htb]
   \centering
   \includegraphics[width=0.9\linewidth]{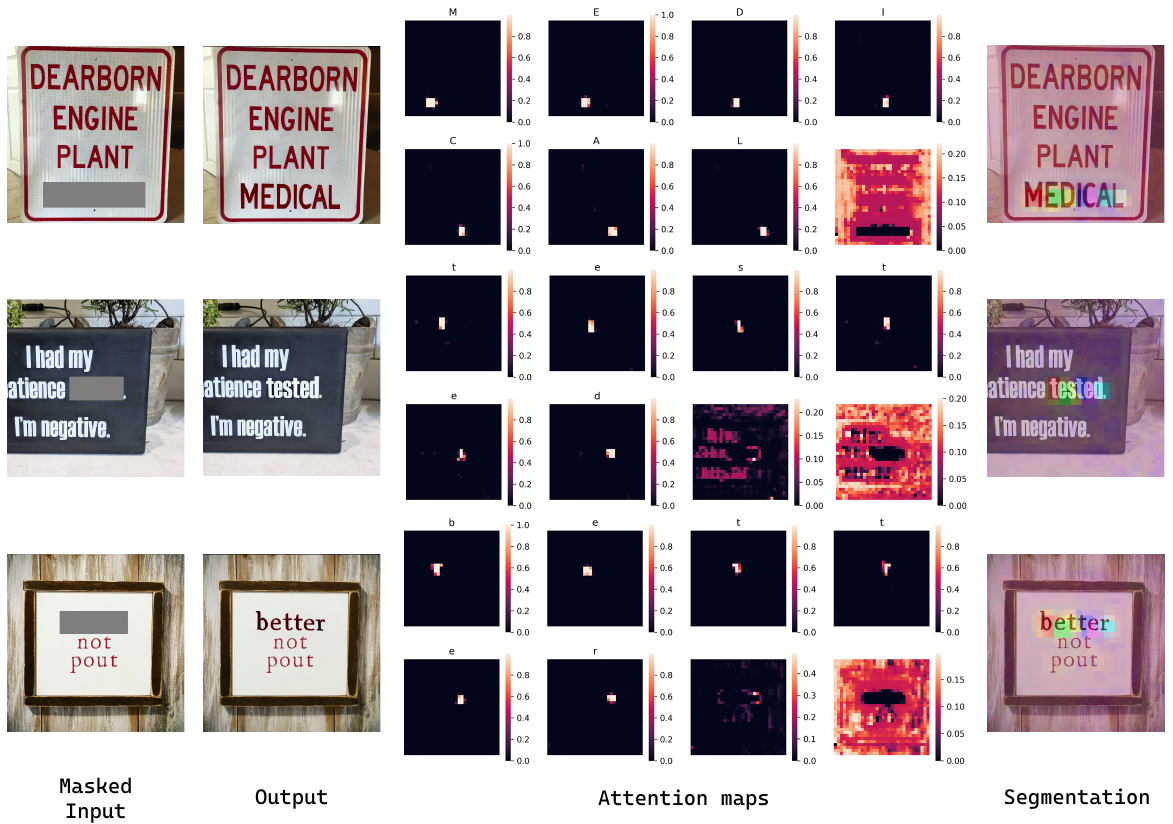}

   \caption{Additional visualization results. The first column is the masked inputs for our UDiffText while the second column shows the outputs. The attention map of each case is extracted from the middle block of the U-Net at intermediate sampling step. The specific token of each attention map is annotated at the top of each map. We up-sample the attention maps to get segmentation maps of the generated images, which are demonstrated at the last column.}
   \label{fig:add_visual}
\end{figure*}

\section{Failure Cases}

Despite its ability to render coherent text in arbitrary given images, our method can still produce unsatisfactory results, including rendering text with distorted characters, repeated characters, incorrect characters and missing characters. These failure cases occur more frequently when the text to be rendered is relatively long or when the masked region is excessively oblique.

\begin{figure*}[htb]
   \centering
   \includegraphics[width=0.9\linewidth]{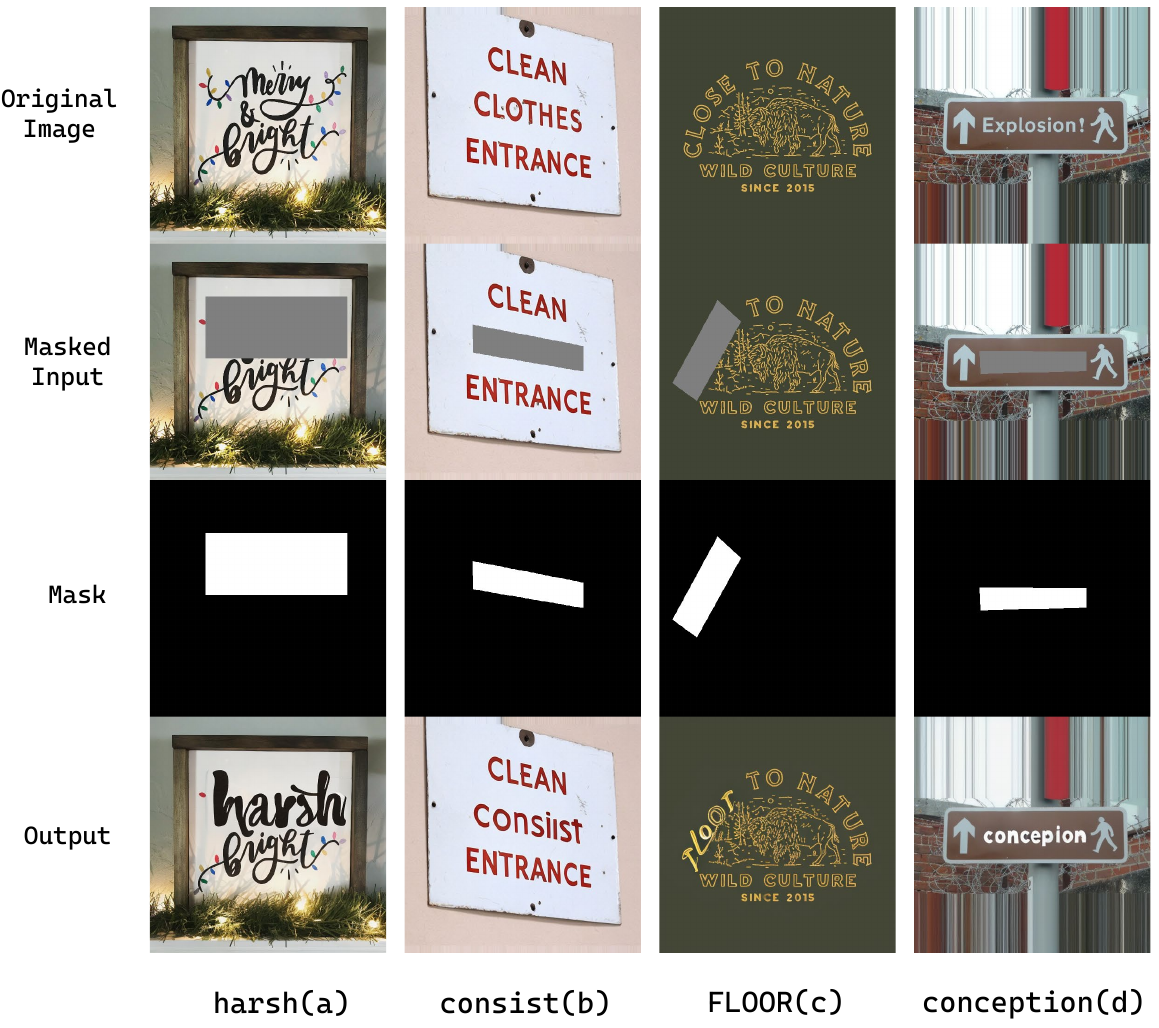}

   \caption{Failure cases. We show some unsatisfactory results of our method, including distorted characters (a), repeated characters (b), wrong characters (c) and missing characters (d). The word to be rendered is annotated at the bottom of each case.}
   \label{fig:add_visual}
\end{figure*}

\end{document}